\documentclass[10pt, conference, compsocconf]{IEEEtran}
\IEEEoverridecommandlockouts
\usepackage{cite}
\usepackage{amsmath,amssymb,amsfonts}
\usepackage{algorithmic}
\usepackage{graphicx}
\usepackage{textcomp}
\usepackage{xcolor}
\usepackage{multirow}
\usepackage{booktabs}
\usepackage[pagebackref,breaklinks,colorlinks]{hyperref}

\newcommand*{\ours}{HAR-GCNN }

\def\BibTeX{{\rm B\kern-.05em{\sc i\kern-.025em b}\kern-.08em
    T\kern-.1667em\lower.7ex\hbox{E}\kern-.125emX}}
\begin{document}

\title{HAR-GCNN: Deep Graph CNNs for Human Activity Recognition From Highly Unlabeled Mobile Sensor Data}

\author{\IEEEauthorblockN{
Abduallah Mohamed\IEEEauthorrefmark{1},
Fernando Lejarza\IEEEauthorrefmark{2},
Stephanie Cahail\IEEEauthorrefmark{3},
Christian Claudel\IEEEauthorrefmark{4},
Edison Thomaz\IEEEauthorrefmark{5}}
\IEEEauthorblockA{The University of Texas at Austin\\
\{\IEEEauthorrefmark{1}abduallah.mohamed,
\IEEEauthorrefmark{2}lejarza,
\IEEEauthorrefmark{3}stephaniecahail,
\IEEEauthorrefmark{4}christian.claudel,
\IEEEauthorrefmark{5}ethomaz\}@utexas.edu}}

\maketitle

\begin{abstract}
The problem of human activity recognition from mobile sensor data applies to multiple domains, such as health monitoring, personal fitness, daily life logging, and senior care. A critical challenge for training human activity recognition models is data quality. Acquiring balanced datasets containing accurate activity labels requires humans to correctly annotate and potentially interfere with the subjects' normal activities in real-time. Despite the likelihood of incorrect annotation or lack thereof, there is often an inherent chronology to human behavior. For example, we take a shower after we exercise. This implicit chronology can be used to learn unknown labels and classify future activities. In this work, we propose HAR-GCCN, a deep graph CNN model that leverages the correlation between chronologically adjacent sensor measurements to predict the correct labels for unclassified activities that have at least one activity label. We propose a new training strategy enforcing that the model predicts the missing activity labels by leveraging the known ones. HAR-GCCN shows superior performance relative to previously used baseline methods, improving classification accuracy by about 25\% and up to 68\% on different datasets. Code is available at \url{https://github.com/abduallahmohamed/HAR-GCNN}.
\end{abstract}


\section{Introduction}
\label{sec_intro}
Human Activity Recognition (HAR) has been an active research field recently, covering a wide range of applications in health monitoring and fitness~\cite{zeng2014convolutional, adaimi2019activelearning}. Technological advances regarding inertial sensors with longer battery life spans and improved computing capabilities have enabled gathering larger volumes of continuous data to be used for HAR~\cite{bulling2014tutorial}.   
Despite these recent advances, ``ground truth annotation'', which involves labeling sensor readings, remains a critical challenge~\cite{bulling2014tutorial}. Such annotation tasks are typically performed manually and occur either in real-time or \textit{post hoc} once the activity has been completed. In the HAR studies, such annotation tasks can be expensive, onerous, prone to human error, and can even condition the user and interfere in the activity itself~\cite{kwon2019handling}. Thus, the training data likely contains a significant amount of unreliable and missing labels. These incorrect or missing labels create gaps in the data which have a detrimental impact on model development and training. Thus, a paramount aspect of models for HAR is their ability to learn in the presence of missing labels, while achieving high accuracy in classifying human activity.
 \begin{figure}[h]
  \includegraphics[width=\columnwidth]{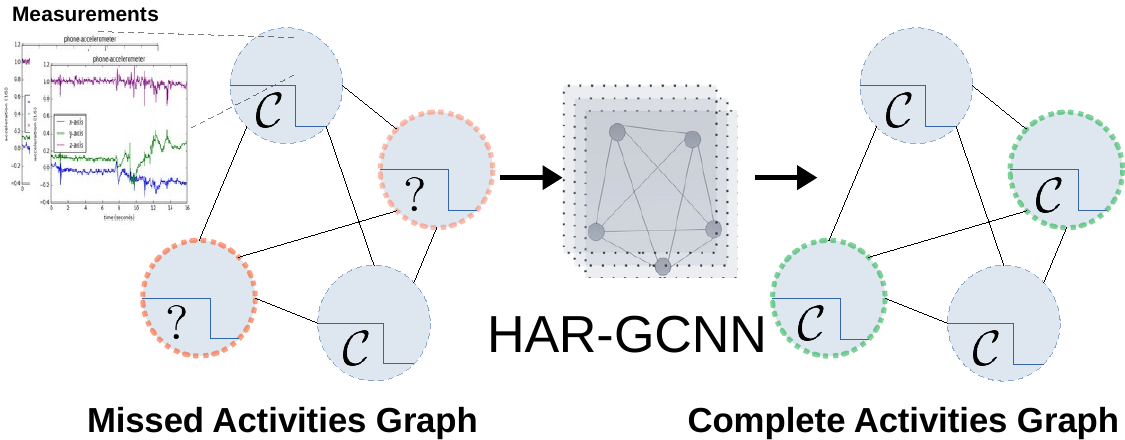}
  \caption{The input to \ours is a graph comprised of partially labelled sensor measurements that are chronologically ordered activities. The model predicts the classes $\mathcal{C}$ of the unlabeled activities. Each input node contains a period of sensor measurements alongside a label or multi-labels. The output is the activity class(es) for each node.}
\label{gr:modelpic}
\end{figure}
A variety of deep and machine learning methods have been previously proposed for single and multi-label classification of human activity~\cite{mantyjarvi2001recognizing,wang2019deep,hammerla2016deep}. The missing label(s) can be predicted based on the readings of multiple sensors such as accelerometer, gyroscope, location, etc~\cite{wang2019deep}. However, fewer approaches leverage the context of ``neighboring'' activities for predicting a missing label which has been shown to improve HAR~\cite{hammerla2016deep,cao2018gchar}. 

We hypothesize that human activities follow a chronological correlation which can provide informational context to improve HAR. The benefit of this hypothesis, if valid, is that one can leverage known or correctly labeled activities to predict the surrounding ones. In order to investigate the validity this assumption, we formulated the HAR problem comprising a sequence of chronologically ordered activities, some of which were correctly labeled while others were not labeled at all. We used deep learning as a data-driven approach to discover this chronological correlation. We first employed Recurrent Neural Nets (RNNs) which is the most straightforward deep model used to handle time series data. Nonetheless, our literature survey revealed that Convolutional Neural Networks (CNNs) can often be more powerful than RNNs in performing HAR~\cite{mohamed2020social}. Yet, both approaches CNNs and RNNS are structurally not geared towards directly leveraging the correlation between the sequential activities. For example, RNNs treat each time step separately by fusing it into the ``neural memory'', which is the only component that partially correlates the data. On a similar note, CNNs only make use of neighbouring activity information based on the chosen kernel size. Based on the shortcomings of these architectures, we posited that graph-based structures more adequately capture the aforementioned features of the problem at hand. Graph representations in HAR allow for modeling each activity as a node, and the graph edges can directly model the relationship between these activities. A suitable tool for learning such graphs is deep Graph CNNs (GCNNs)~\cite{kipf2016semi}. GCNNs behave as CNNs but weight the nodes based on the value of the edges to exploit the correlation between sequential activities.

In this work we show that the chronological correlation indeed exists by experimenting with two commonly employed HAR datasets. One was collected in the wild, while the other was collected in a scripted manner. Our results show that the proposed models benefit from this correlation and use it to predict the neighbouring missing activities, improving performance relative to RNNs and CNNs benchmarks. In the following sections, we discuss prior works, as well as the datasets utilized for training and validation. We provide details of our formulation including the graph structure used to model human activities. Then, we introduce our proposed HAR-GCNN, highlighting details of its architecture and implementation, and define the comparison baselines used to benchmark our approach. In the experiments section, we evaluate our model's performance relative to other deep neural network architectures employing widely used HAR datasets.  
 
\section{Related Work}
\label{sec_rwork}
A large number of traditional supervised machine learning techniques have been developed in the literature for HAR from mobile sensor data. Such models include, for example, logistic regression, $k$-nearest neighbors, decision trees, and multi-layer perceptrons (MLP)~\cite{mantyjarvi2001recognizing,kwapisz2011activity,pirttikangas2006feature}. These models typically exhibit good performance when trained on controlled datasets that are fully and accurately labeled, which may not be the case for in-the-wild environments~\cite{vaizman2018context}. Furthermore, these models often require substantial feature extraction which can be time-consuming and rely heavily on domain knowledge~\cite{wang2019deep}. 

Addressing these shortcomings, a variety of prior works have explored unsupervised and semi-supervised learning techniques on human activity data. For example,~\cite{Varamin2018deepautoset} developed a network of convolutional autoencoders on an unlabeled dataset to extract useful features, which were then used to complement their supervised learning mechanism.~\cite{Varamin2018deepautoset} proposed Deep Auto-Set, a deep learning classification model trained on raw multi-modal sensory segments. Furthermore,~\cite{adaimi2019activelearning} implemented semi-supervised, active learning techniques operated both online and offline.~\cite{adaimi2019activelearning} demonstrated superior performance relative to fully supervised approaches on HAR datasets where ground truth labels are scarce. More specifically,~\cite{adaimi2019activelearning} evaluated the performance of different pool-based and stream-based active learning frameworks on various datasets relevant to human activity~\cite{vaizman2017recognizing} and contrasted classification accuracy against classical machine learning approaches. 

Besides these models, a great number of deep learning frameworks have been proposed for HAR~\cite{wang2019deep}. For example, different CNN architectures~\cite{grzeszick2017deep, moya2018convolutional, cruciani2020feature} resulted in significantly higher accuracy than prior machine learning techniques applied on HAR. Further,~\cite{munzner2017cnn} explored normalization and sensor data fusion using CNNs and demonstrated that these techniques can further improve the performance of deep learning models.~\cite{hammerla2016deep} investigated deep convolutional recurrent models including deep feed-forward neural networks (DNN), CNNs, and different variants of Long Short-Term Memory  (LSTM) cells. Their results show that recurrent networks significantly outperform convolutional networks on activities that are short in duration but follow a natural chronology. Such performance gains are likely because recurrent models can contextualize to improve recognition.~\cite{guan2017ensembles} reported improved HAR using ensembles of deep LSTM learners relative to previously reported recurrent neural networks.~\cite{inoue2018deep} developed a deep RNN performing extensive hyper-parameters optimization and showed superior performance compared to other traditional machine learning models. 

Several variants of graph-based models have been recently proposed leveraging spatial and temporal properties in the data for collective HAR (i.e., predicting the activity of an entire group as opposed to a single subject). Numerous of these models have been reported for computer vision applications to (deeply) learn the interactions between individual participants for a given sequence of video clips and to predict the overall activity or outcome of a group of people~\cite{lu2019gaim, zhang2020temporal, wu2019learning, singh2017graph, tang2019coherence}. Similarly,~\cite{stikic2009multi} proposed a context-aware, semi-supervised graph propagation algorithm with a Support Vector Machine (SVM) classifier to address individual HAR. The results in~\cite{stikic2009multi} suggest that exploiting contextual data from neighboring activities (i.e., nodes) can result in improved performance even relative to fully supervised approaches, likely by discriminating outliers and erroneously labeled data.

GCNNs have recently gained significant popularity for various applications of semi-supervised learning~\cite{kipf2016semi, zhou2018graph}. GCNNs leverage dependencies between the features and labels of nodes in a given graph resulting in improved predictive performance, particularly when a significant number of the training labels are missing. Considering these desirable characteristics, as well as the aforementioned advantages of models that can contextualize, GCNNs are an appealing modeling strategy that to the best of our knowledge has not been previously reported for HAR. In light of this, the main contributions of this work are:
\begin{itemize}
    \item A formulation for HAR exploiting the chronological context of the activities embedded within a graph structure.
    \item A novel mechanism that trains \ours to learn to predict the missing labels in the input graph with high accuracy. 
    \item Extensive computational experiments evaluating the effect of the percentage of missing labels, as well as the effect of the number of activities on the prediction accuracy. Experiments performed on Extra-Sensory~\cite{vaizman2017recognizing} and PAMAP~\cite{reiss2012introducing} datasets, which are among the most commonly used in the HAR literature.
    \item Benchmarks of our proposed approach against previously reported deep architectures including CNN and LSTM models.  
\end{itemize}

\section{Problem Formulation \& Datasets}
Given a set of time series sensory measurements $\mathcal{F} =\{f_t| t \in T\}$ sampled over a window of time steps $T$, it is of interest to learn the classes of activities $\mathcal{C} = \{c_t| t \in T\}$ associated with each of such measurements. The measurements are collected from a variety of sensors such as accelerometers, gyrometers, etc. Each interval of measurements is associated with multiple labels corresponding to different activities (i.e., multi-label classification). Our problem formulation uses a set of prior and posterior measurements with known activities to predict the multi-label or single label classes of those unknown activities. In other words, using ${(f_{t-m},c_{t-m}), ... ,(f_{t-1},c_{t-1}), (f_{t+1},c_{t+1}), ... , (f_{t+m},c_{t+m})}$ and $f_t$ to predict $c_t$, where $m$ is the number of neighbour activities to be used. In this way, we consider a sequence of activities represented by sensor measurements over a time horizon including the associated labels, whether they are known or not, and predict the class of the unknown labels exploiting the observed sequential order. One important remark is that the datasets under consideration are recorded \textit{in-the-wild}. That is, an in-the-wild dataset contains sensor readings from the users who were not previously instructed to perform a given set of activities (which is typically the case in more controlled environments). In-the-wild settings in turn reduce the bias within the collected data and results in a more natural chronology of activities from which our proposed model can learn. The Extra-Sensory~\cite{vaizman2017recognizing} dataset is an example of such in-the-wild measurements. The dataset contains over 300,000 minutes of labeled sensor recordings from smartphones and smartwatches worn by a total of 60 study participants. The behavioural activities in this dataset can be categorized by 51 non-exclusive labels such as sitting, sleeping, strolling, cooking, etc. For this type of the data the HAR task at hand is a \textit{mutli-label classification problem}. Further, the dataset contains 224 different raw features obtained from mobile sensor readings recorded over 20 seconds window every one minute. To further validate the advantages of the framework proposed herein, we also conduct experiments using the PAMAP dataset~\cite{reiss2012introducing}. The PAMAP dataset while collected in a more controlled environment provides a useful instance to validate our framework, particularly when labels are artificially hidden during training. The PAMAP dataset consists of data from 9 subjects, wearing 3 inertial measurement units and a heart rate monitor and performing 12 exclusive activities (i.e., the HAR task is a single-label classification problem) with 52 raw features per instance. The PAMAP dataset is being used herein as an example of a scripted dataset, which likely means that the chronological order of the recorded activities indeed contains some amount of bias and which we further emphasize in the numerical experiments section.  
\textbf{Training/ Testing details: } Both datasets were split into training and test sets with a ratio of 2:1. The sequence of the recorded activities was kept and no randomization was used, which is crucial for learning from the natural sequential order of activities. The Extra-Sensory dataset serves to show that the activities can be predicted with higher accuracy by exploiting such implicit order. Conversely, the PAMAP dataset is employed to show that if the sequence of activities is scripted \textit{a priori}, \ours{} results in almost perfect classification performance as it can quickly learn the underlying script. We note that no test data was used to train the proposed model, and the data sets were kept separate to prevent any data leaking.

\section{\ours Method Description}
\begin{figure}[ht]
\begin{center}
\includegraphics[width=0.8\columnwidth]{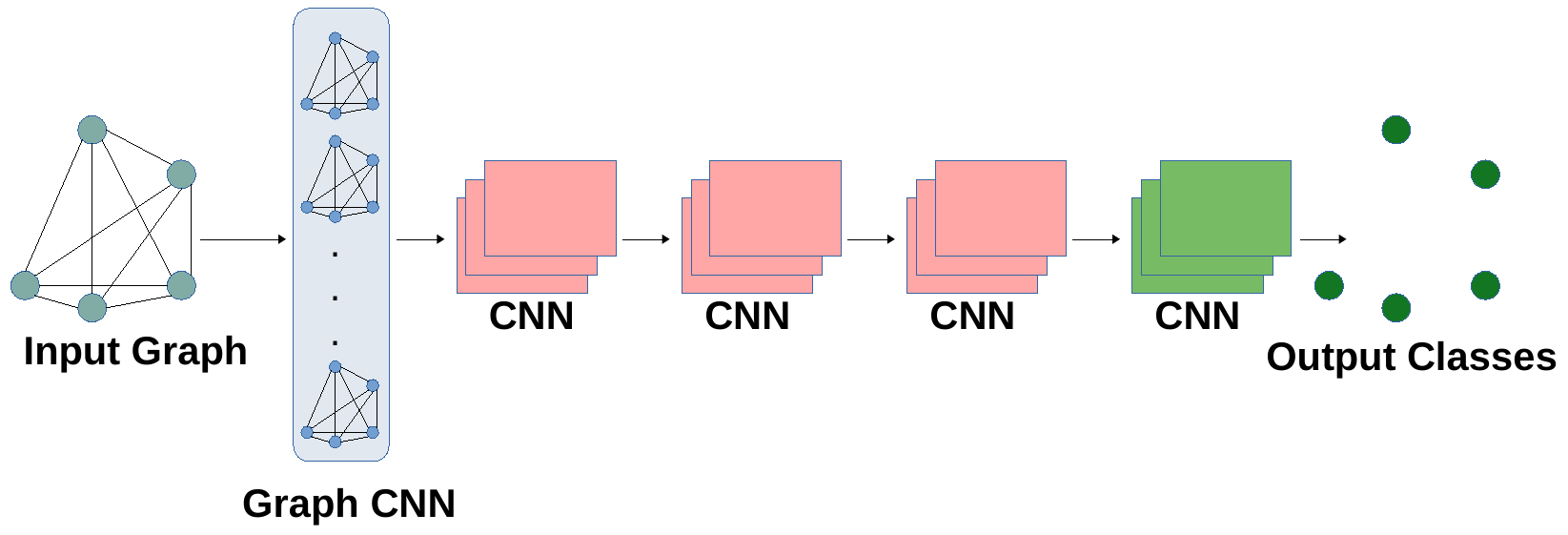}
\end{center}
   \caption{\ours model description. The CNN is just a single layer convolution with an activation function.}
\label{gr:actmodel}
\end{figure}

\textbf{Constructing the activity graph: }We model the set of activities as a graph, which is defined as $\mathcal{G}=(\mathcal{V}, \mathcal{E})$ where $\mathcal{V}=\{v| v \in \mathcal{V}\}$ is a set of vertices. Each vertex $v = [f_t,c_t]$ contains both measurements over a time window and the associated multi-label class, with the exception of the nodes whose activity label is missing and we thus want to predict, for which $c$ is set to zero. The graph edges are fully connected, $\mathcal{E}=\{e_{ij}| e \in \mathcal{E}; i,j \in |V|\}$, where $\mathcal{E}$ is the set of all the graph edges. The weight of these edges is set to one, so that predicting the activity of all nodes is equally important. Our proposed model allows using any number of nodes during training. For example, the graph can consist of three nodes with one of them missing its label. In the experiments section we study the effect of the number of input labelled neighbour nodes on predicting a missing activity class. Further, we note that framework allows for predicting more than one missing label, which we explore in our numerical experiments.

\textbf{Model Description: }First, the \textbf{GCNN layer} takes as input the aforementioned activity graph $\mathcal{G}$. The GCNN model has a layer-wise structure defined as $\text{GCNN}(\mathcal{V}^{(l)}, A) = \sigma\left( A_{\text{norm}}\mathcal{V}^{(l)}W^{(l)}\right)$ Where $A$ is the adjacency matrix that defines the edges of the activity graph and $\sigma(\cdot)$ denotes an activation function. The GCNN operates as an ordinary CNN except that it weights the kernel by the value of the normalized adjacency matrix $A_{\text{norm}}$ defined as: 
\[A_{\text{norm}}= I- \hat{D}^{-\frac{1}{2}} (A+I) \hat{D}^{-\frac{1}{2}}\] where $\hat{D}$ is the degree node matrix of $A+I$, and $I$ is the identity matrix. This normalization approach was introduced by~\cite{mohamed2020social}. More in-depth information about graph CNNs can be found in~\cite{kipf2016semi}. The output of the GCNN is a graph embedding that represents the whole information of the sensor measurements and their corresponding known labels. 

The second step in \ours is the \textbf{CNN output layers}, for which we use a sequence of CNNs (single layer convolutions with an activation function). This architecture was chosen mainly because  performance can deteriorate when the depth of the GCNNS increases \cite{9010334}. That is, it would be inefficient to use a sequence of graph CNNs to predict each activity label. The output of these CNN layers is directly regressed against the proper loss function (Binary Cross Entropy or Cross-Entropy) to predict the labels $\mathcal{C}$. Figure~\ref{gr:actmodel} describes the distinct steps included in our architecture.

\textbf{Implementation\footnote{Code: https://github.com/abduallahmohamed/HAR-GCNN.git} Details: }We employ the same GCNN implementation as the one in the pioneering work of~\cite{kipf2016semi}. The chosen implementation is easier to interpret and deploy, relative to similar architectures available in PyTorch. The model consists of a single GCNN layer that outputs a graph embedding. Then, three CNN layers are used to process the embedding and a final CNN layer is included with a sigmoid (softmax for PAMAP) activation function. We used PReLU~\cite{DBLP:journals/corr/HeZR015} activation functions for intermediate layers. The total model parameters size is ${\sim}15k$ for the Extra-Sensory dataset and ${\sim}5k$ for the PAMAP dataset. 
\section{Baseline Methods}
\begin{figure}[t]
\begin{center}
\includegraphics[width=0.8\linewidth]{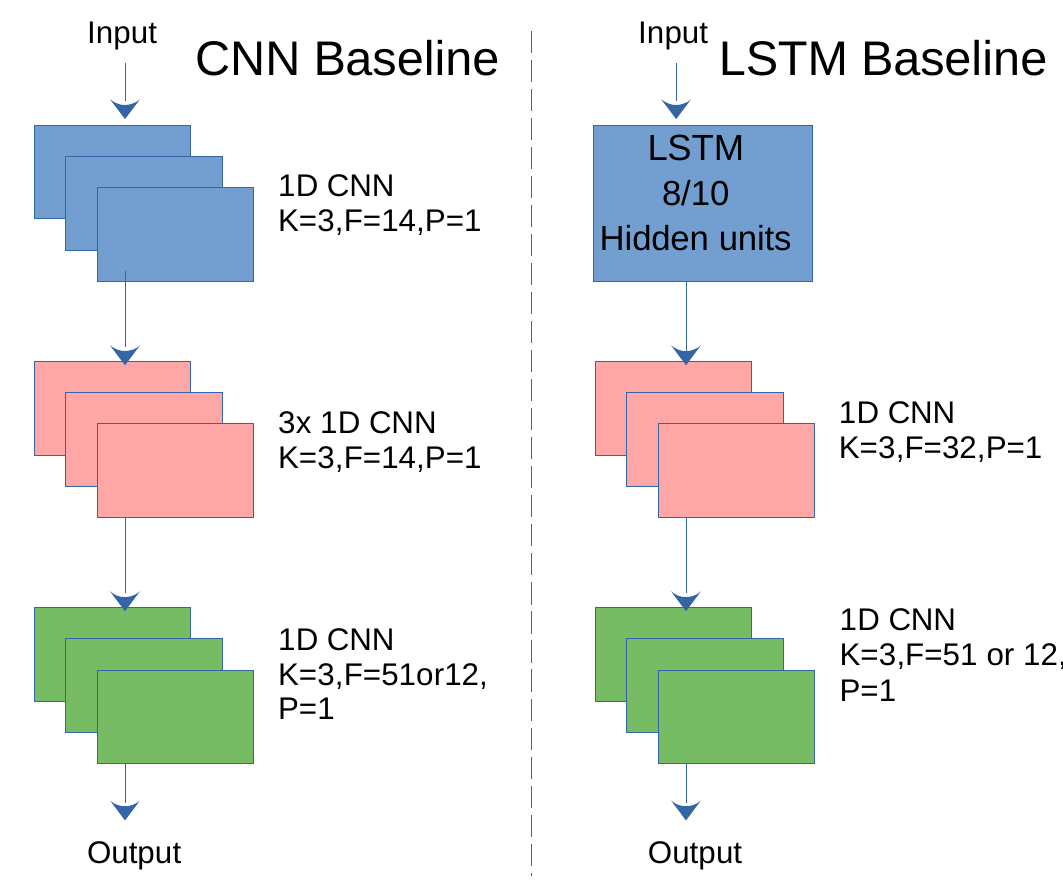}
\end{center}
   \caption{CNN and LSTM baselines. Both baselines share the same depth and the same number of parameters as in our model. $K$ is the kernel size, $P$ is the padding size and $F$ is the features. The output layer changes its features based on the number of predicted classes. All CNNs are single layer convolution followed by an activation function.}
\label{gr:actbaslines}
\end{figure}

In order to benchmark the performance of our approach, we compared it to the aforementioned, more commonly encountered deep architectures. Mainly, we contrast \ours to CNNs~\cite{grzeszick2017deep, moya2018convolutional,cruciani2020feature} and LSTMs~\cite{hammerla2016deep,sainath2015convolutional} that were previously used for HAR. \textbf{CNN baseline} The CNN baseline, shown in Figure~\ref{gr:actbaslines},  is a sequence of five CNN layers that have the same depth as \ours model. The reason for choosing this architecture is that $\mathcal{V}$ represents a set of an arbitrary number of activities, and a CNN can learn a kernel that is agnostic to the width and height of the input. The total parameters size is also ${\sim}15k$ for the Extra-Sensory and ${\sim}5k$ for the PAMAP datasets, thus allowing for a fair comparison with HAR-GCNN. The input to this base model is the $\mathcal{V}$ itself, which can be interpreted as an image of width equal to the sensor readings, and of height equal to the number of nodes or activities in the graph. All layers used PReLU~\cite{DBLP:journals/corr/HeZR015} activation function except the last layer with a sigmoid (softmax for PAMAP) function. The intermediate three layers use a residual connection to improve performance during training. \textbf{LSTM baseline} The inputs to the LSTM, shown in Figure~\ref{gr:actbaslines}, are the graph nodes treated as a sequence of time steps. The model parameter size is also ${\sim}15k$ and ${\sim}5k$ for both Extra-Sensory and PAMAP respectively, same as \ours model and CNN base model. The 1D-CNN layers use PReLU activation function except for the last layer which uses a sigmoid (softmax for PAMAP) activation function. 
\section{Training Mechanism}
\label{sec:trainmecha}
\begin{figure*}[h]
\scriptsize
\begin{center}
\includegraphics[width=\linewidth]{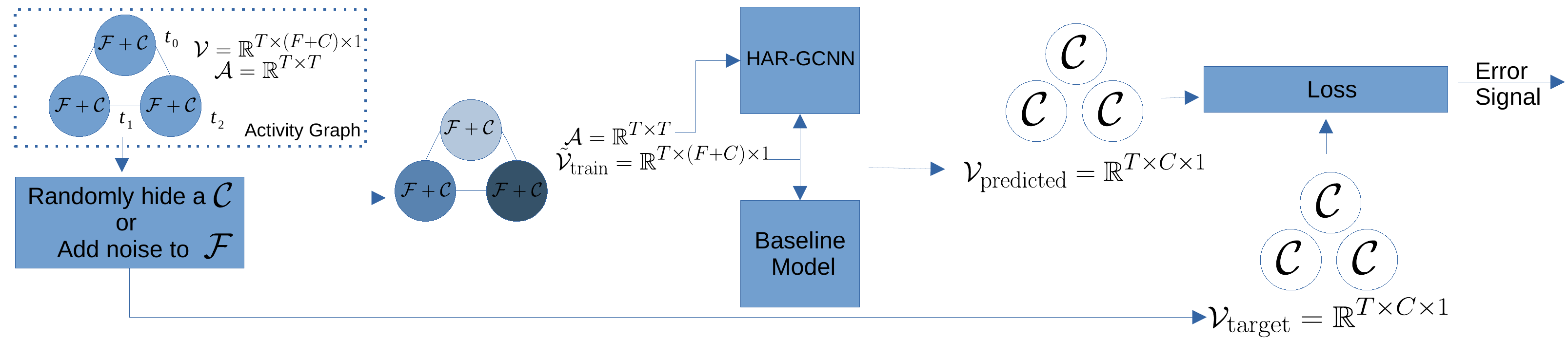}
\end{center}
\caption{\ours data flow diagram for model training as described in ~\ref{sec:trainmecha}. $\mathcal{F}$ represents the sensor measurements with dimension $F={224,52}$ for the Extra-Sensory dataset and PAMAP dataset, respectively. $\mathcal{C}$ is the set of multi-label activity classes with dimensions $C={51,12}$ for the Extra-Sensory dataset and PAMAP dataset, respectively. $T$ is the number of time steps. $\mathcal{A}$ stands for adjacency matrix, $\mathcal{V}$ stands for the graph vertices.}
\label{gr:actdataflow}
\end{figure*}

Besides keeping the number of parameters consistent across models, we trained all models using the same  settings. We used Binary Cross Entropy (BCE) as the loss function in the mutli-label case of the Extra-Sensory dataset, which defined as: 
$\text{BCE loss}_j = - \left[ \text{class}_j \cdot \log x_j + (1 -\text{class}_j ) \cdot \log (1 - x_j) \right]$
For the single-label PAMAP dataset we used the Crossy-Entropy(CE) loss, which is defined as:
$ \text{CE loss}(x, class) = -\log\left(\frac{\exp(x[class])}{\sum_j \exp(x[j])}\right)$
where $j$ is the class number and $x$ is the predicted class. We artificially imposed missing labels in both datasets to simulate the case of missing labels. Our training framework randomly hides a percentage of the known activity labels $\mathcal{C}$ within the input graph $\mathcal{G}$, which serve as the missing labels we want to predict. Moreover to make the model generalize better, we add random noise (Gaussian with 0 mean and 1 standard deviation) to disturb the measurements vector $\mathcal{F}$, which helps generalize the performance of \ours for unseen data while also accounting for possible measurement errors in the raw sensor data. In all of our experiments, we used a 50\% probability for either hiding a label or adding noise to the measurements. Further, we restricted the percentage of hidden nodes to be no greater than 66\%, so that we are guaranteed that at least 33\% of the nodes have their original labels. The disturbed data were generated once and were used in all of our experiments across different models for consistency. All of our results are reported on the withheld test set. We report the mean of three repetitions of our experiments, which were conducted with a controlled random seed to ensure consistent settings. Figure~\ref{gr:actdataflow} shows the data-flow and the dimension of the data alongside the training mechanism. 

\section{Experiments \& Discussion}
\begin{table*}[ht]
\centering
\scriptsize
\caption{F-1 score over accuracy of \ours versus baseline models (the higher the better). The Extra-Sensory results are $F_{\text{1 macro}}$ / $\text{Acc}_{\text{mean}}$, and PAMAP results are $F_1/$Acc. 3, 5, 10 and 25 chronologically ordered activities were used over a range of missing labels on accuracy over F-1 score. The average of three runs is reported, and bolded numbers show the best.}
\label{tab:nodes}
\begin{tabular}{|l|l|l|l|l||l|l|l|} 
\toprule
\multirow{2}{*}{\# of Activities} & \multirow{2}{*}{\% missing labels} & \multicolumn{3}{l||}{\hfil Extra-Sensory Dataset}               & \multicolumn{3}{l|}{\hfil PAMAP Dataset}                      \\ 
\cline{3-8}
                      &                                   & \textbf{\ours }          & CNN           & LSTM          & \textbf{\ours }          & CNN           & LSTM           \\ 
\hline
\multirow{3}{*}{\hfil 3}    & \hfil 33\%                                & \textbf{0.781 / 99.52} & 0.621 / 99.23 & 0.464 / 98.43 & \textbf{0.999 / 99.92} & 0.975 / 97.52 & 0.973 / 97.26  \\
                      & \hfil 66\%                          & \textbf{0.792 / 99.52} & 0.531 / 98.69 & 0.377 / 97.95 & \textbf{0.999 / 99.94} & 0.902 / 90.22 & 0.903 / 90.26  \\
\hline
\multirow{3}{*}{\hfil 5}    & \hfil 33\%                                & \textbf{0.814 / 99.41} & 0.715 / 99.41 & 0.504 / 98.67 & 0.998 / 99.76 & \textbf{1.000 / 99.96} & 0.990 / 99.05  \\
                      & \hfil 66\%                          & \textbf{0.880 / 99.58} & 0.659 / 99.25 & 0.459 / 98.37 & \textbf{1.000 / 99.98} & 0.996 / 99.57 & 0.950 / 95.05  \\
\hline
\multirow{3}{*}{\hfil 10}   & \hfil 33\%                                 & \textbf{0.813 / 99.14} & 0.744 / 99.46 & 0.522 / 98.64 & \textbf{0.999 / 99.91} & 0.998 / 99.81 & 0.997 / 99.65  \\
                      & \hfil 66\%                                 & \textbf{0.868 / 99.52} & 0.691 / 99.28 & 0.483 / 98.37 & \textbf{1.000 / 99.98} & 0.997 / 99.75 & 0.969 / 96.87  \\
\hline
\multirow{3}{*}{\hfil 25}   & \hfil 30\%                                 & 0.748 / 98.76 & \textbf{0.784 / 99.50} & 0.532 / 98.73 & 0.998 / 99.79 & \textbf{1.000 / 99.99} & 0.998 / 99.75  \\
                      & \hfil 66\%                                 & \textbf{0.838 / 99.41} & 0.724 / 99.23 & 0.494 / 98.46 & \textbf{1.000 / 99.99} & 0.997 / 99.68 & 0.981 / 98.10  \\
\hline
\end{tabular}

\end{table*}
We considered two settings for the amount of missing class labels $\mathcal{C}$: 33\% and 66\% in the input graph $\mathcal{G}$. For example, the case of 66\% missing labels corresponds to $\frac{2}{3}$ of the measurements $\mathcal{F}$ being known but without labels $\mathcal{C}$, which we seek to predict. This method evaluates the model's ability to generalize to unseen instances because the models were trained under different settings as described in Section~\ref{sec:trainmecha}. For the Extra-Sensory dataset, we used the macro $F_1$ score which is defined as: $F_{\text{1 macro}} = \frac{1}{N} \sum_{n \in N} F_1^n$ where $N$ is the total number of classes. The mean accuracy is defined as:  $\text{Acc}_{\text{mean}} = \frac{1}{N} \sum_{n \in N} \text{Acc}_n$. For the PAMAP dataset we use the ordinary F1 and accuracy scores.


\subsection{Performance of \ours Against Baselines}

Table~\ref{tab:nodes} shows the main results comparing the performance of \ours versus the baseline models. We explain the main results by focusing on the three activities case (i.e., first row of Table~\ref{tab:nodes}) but note that the observations generalize for all other instances in Table~\ref{tab:nodes}. In the three activities case, 33\% missing labels imply that only one out of the three activities does not have a label. An important result is that \ours, in the case of 33\% or 66\% missing labels, outperforms both CNN and LSTM baselines. The performance improvement is about 25\% and 68\% relative to CNN and LSTM models respectively for the Extra-Sensory dataset. For the PAMAP dataset \ours has a $\sim$2\%  higher F-1 score than both CNN and LSTM, and almost reaching the upper performance bound of 1.00. We posit that the observed performance improvements stem from the proposed graph formulation, in which the graph edges can more deeply capture the relationship between labels and their features. Therefore, if at least one activity is correctly classified, then the remaining nodes will have a higher likelihood of being correctly labeled as well. We note that the CNN baseline significantly outperforms the LSTM model across the two datasets considered. CNN-based approaches appear to be more suitable to this type of problem, as they do not accumulate errors sequentially while classifying the measurements and have a more global view of the entire state. Unlike in recurrent nature of methods like LSTM, the error in previous predictions propagates to the future ones. These findings are in line with prior works~\cite{mohamed2020social}.

\subsection{Effect of The Cardinality of The Graph Nodes }

We also explored the effect of the number of chronologically ordered activities on the classification accuracy. From Table~\ref{tab:nodes} it is noticeable that the results discussed in the previous section are consistent for an increasing number of activities in the input graph. From the results obtained using the Extra-Sensory dataset, the performance of all base models improves as more graph nodes are considered. The performance of \ours is still consistently better than the base models and does not improve substantially when the number of activities is increased from 5 to 25. These results imply that \ours learns a proper representation that has a constant performance irrelevant from the number of activities. The results obtained using the PAMAP dataset in Table~\ref{tab:nodes} show that the performance across all models (\ours and baseline models) does not change significantly with the graph cardinality. This constant performance can be attributed to the PAMAP dataset, which was collected from users being asked to perform tasks in a pre-scripted manner. Models trained on this dataset might readily infer such pre-scripted sequence of activities and thus resulting in very high F-1 score and accuracy.

\section{Conclusion}
\label{sec:conclusion}
In this paper, we presented HAR-GCNN, a deep graph model to predict missing activity labels leveraging the context of chronological sequences of these activities. We proposed an approach for modelling the activities as nodes in a fully connected graph. We introduced a new training mechanism that enables the model to learn the implied natural chronological order of human activities. To gain a quantitative understanding of our proposed strategy, we benchmarked our design against baseline deep models sharing the similar structural properties as HAR-GCNN. Our results indicate that \ours has superior performance in terms of F-1 score and accuracy compared to the chosen baselines. Further, our results suggest that learning a chronology of activities to predict the missing label is likely a key driver for such performance improvements. Our experimental results reveal that \ours has a stable performance, independently from the number of chronologically ordered activities considered within the input graph.



\bibliographystyle{IEEEtran}
\bibliography{actgraphbib}

\begin{thebibliography}{10}
\providecommand{\url}[1]{#1}
\csname url@samestyle\endcsname
\providecommand{\newblock}{\relax}
\providecommand{\bibinfo}[2]{#2}
\providecommand{\BIBentrySTDinterwordspacing}{\spaceskip=0pt\relax}
\providecommand{\BIBentryALTinterwordstretchfactor}{4}
\providecommand{\BIBentryALTinterwordspacing}{\spaceskip=\fontdimen2\font plus
\BIBentryALTinterwordstretchfactor\fontdimen3\font minus
  \fontdimen4\font\relax}
\providecommand{\BIBforeignlanguage}[2]{{%
\expandafter\ifx\csname l@#1\endcsname\relax
\typeout{** WARNING: IEEEtran.bst: No hyphenation pattern has been}%
\typeout{** loaded for the language `#1'. Using the pattern for}%
\typeout{** the default language instead.}%
\else
\language=\csname l@#1\endcsname
\fi
#2}}
\providecommand{\BIBdecl}{\relax}
\BIBdecl

\bibitem{zeng2014convolutional}
M.~Zeng, L.~T. Nguyen, B.~Yu, O.~J. Mengshoel, J.~Zhu, P.~Wu, and J.~Zhang,
  ``Convolutional neural networks for human activity recognition using mobile
  sensors,'' in \emph{6th International Conference on Mobile Computing,
  Applications and Services}.\hskip 1em plus 0.5em minus 0.4em\relax IEEE,
  2014, pp. 197--205.

\bibitem{adaimi2019activelearning}
\BIBentryALTinterwordspacing
R.~Adaimi and E.~Thomaz, ``Leveraging active learning and conditional mutual
  information to minimize data annotation in human activity recognition,''
  \emph{Proc. ACM Interact. Mob. Wearable Ubiquitous Technol.}, vol.~3, no.~3,
  Sep. 2019. [Online]. Available:
  \url{https://doi-org.ezproxy.lib.utexas.edu/10.1145/3351228}
\BIBentrySTDinterwordspacing

\bibitem{bulling2014tutorial}
A.~Bulling, U.~Blanke, and B.~Schiele, ``A tutorial on human activity
  recognition using body-worn inertial sensors,'' \emph{ACM Computing Surveys
  (CSUR)}, vol.~46, no.~3, pp. 1--33, 2014.

\bibitem{kwon2019handling}
H.~Kwon, G.~D. Abowd, and T.~Pl{\"o}tz, ``Handling annotation uncertainty in
  human activity recognition,'' in \emph{Proceedings of the 23rd International
  Symposium on Wearable Computers}, 2019, pp. 109--117.

\bibitem{mantyjarvi2001recognizing}
J.~Mantyjarvi, J.~Himberg, and T.~Seppanen, ``Recognizing human motion with
  multiple acceleration sensors,'' in \emph{2001 IEEE International Conference
  on Systems, Man and Cybernetics. e-Systems and e-Man for Cybernetics in
  Cyberspace (Cat. No. 01CH37236)}, vol.~2.\hskip 1em plus 0.5em minus
  0.4em\relax IEEE, 2001, pp. 747--752.

\bibitem{wang2019deep}
J.~Wang, Y.~Chen, S.~Hao, X.~Peng, and L.~Hu, ``Deep learning for sensor-based
  activity recognition: A survey,'' \emph{Pattern Recognition Letters}, vol.
  119, pp. 3--11, 2019.

\bibitem{hammerla2016deep}
N.~Y. Hammerla, S.~Halloran, and T.~Pl{\"o}tz, ``Deep, convolutional, and
  recurrent models for human activity recognition using wearables,''
  \emph{arXiv preprint arXiv:1604.08880}, 2016.

\bibitem{cao2018gchar}
L.~Cao, Y.~Wang, B.~Zhang, Q.~Jin, and A.~V. Vasilakos, ``Gchar: An efficient
  group-based context—aware human activity recognition on smartphone,''
  \emph{Journal of Parallel and Distributed Computing}, vol. 118, pp. 67--80,
  2018.

\bibitem{mohamed2020social}
A.~Mohamed, K.~Qian, M.~Elhoseiny, and C.~Claudel, ``Social-stgcnn: A social
  spatio-temporal graph convolutional neural network for human trajectory
  prediction,'' in \emph{Proceedings of the IEEE/CVF Conference on Computer
  Vision and Pattern Recognition}, 2020, pp. 14\,424--14\,432.

\bibitem{kipf2016semi}
T.~N. Kipf and M.~Welling, ``Semi-supervised classification with graph
  convolutional networks,'' \emph{arXiv preprint arXiv:1609.02907}, 2016.

\bibitem{kwapisz2011activity}
J.~R. Kwapisz, G.~M. Weiss, and S.~A. Moore, ``Activity recognition using cell
  phone accelerometers,'' \emph{ACM SigKDD Explorations Newsletter}, vol.~12,
  no.~2, pp. 74--82, 2011.

\bibitem{pirttikangas2006feature}
S.~Pirttikangas, K.~Fujinami, and T.~Nakajima, ``Feature selection and activity
  recognition from wearable sensors,'' in \emph{International symposium on
  ubiquitious computing systems}.\hskip 1em plus 0.5em minus 0.4em\relax
  Springer, 2006, pp. 516--527.

\bibitem{vaizman2018context}
Y.~Vaizman, N.~Weibel, and G.~Lanckriet, ``Context recognition in-the-wild:
  Unified model for multi-modal sensors and multi-label classification,''
  \emph{Proceedings of the ACM on Interactive, Mobile, Wearable and Ubiquitous
  Technologies}, vol.~1, no.~4, pp. 1--22, 2018.

\bibitem{Varamin2018deepautoset}
\BIBentryALTinterwordspacing
A.~A. Varamin, E.~Abbasnejad, Q.~Shi, D.~C. Ranasinghe, and H.~Rezatofighi,
  ``Deep auto-set: A deep auto-encoder-set network for activity recognition
  using wearables,'' in \emph{Proceedings of the 15th EAI International
  Conference on Mobile and Ubiquitous Systems: Computing, Networking and
  Services}, ser. MobiQuitous ’18.\hskip 1em plus 0.5em minus 0.4em\relax New
  York, NY, USA: Association for Computing Machinery, 2018, p. 246–253.
  [Online]. Available: \url{https://doi.org/10.1145/3286978.3287024}
\BIBentrySTDinterwordspacing

\bibitem{vaizman2017recognizing}
Y.~Vaizman, K.~Ellis, and G.~Lanckriet, ``Recognizing detailed human context in
  the wild from smartphones and smartwatches,'' \emph{IEEE Pervasive
  Computing}, vol.~16, no.~4, pp. 62--74, 2017.

\bibitem{grzeszick2017deep}
R.~Grzeszick, J.~M. Lenk, F.~M. Rueda, G.~A. Fink, S.~Feldhorst, and M.~ten
  Hompel, ``Deep neural network based human activity recognition for the order
  picking process,'' in \emph{Proceedings of the 4th international Workshop on
  Sensor-based Activity Recognition and Interaction}, 2017, pp. 1--6.

\bibitem{moya2018convolutional}
F.~Moya~Rueda, R.~Grzeszick, G.~A. Fink, S.~Feldhorst, and M.~Ten~Hompel,
  ``Convolutional neural networks for human activity recognition using
  body-worn sensors,'' in \emph{Informatics}, vol.~5, no.~2.\hskip 1em plus
  0.5em minus 0.4em\relax Multidisciplinary Digital Publishing Institute, 2018,
  p.~26.

\bibitem{cruciani2020feature}
F.~Cruciani, A.~Vafeiadis, C.~Nugent, I.~Cleland, P.~McCullagh, K.~Votis,
  D.~Giakoumis, D.~Tzovaras, L.~Chen, and R.~Hamzaoui, ``Feature learning for
  human activity recognition using convolutional neural networks,'' \emph{CCF
  Transactions on Pervasive Computing and Interaction}, vol.~2, no.~1, pp.
  18--32, 2020.

\bibitem{munzner2017cnn}
S.~M{\"u}nzner, P.~Schmidt, A.~Reiss, M.~Hanselmann, R.~Stiefelhagen, and
  R.~D{\"u}richen, ``Cnn-based sensor fusion techniques for multimodal human
  activity recognition,'' in \emph{Proceedings of the 2017 ACM International
  Symposium on Wearable Computers}, 2017, pp. 158--165.

\bibitem{guan2017ensembles}
Y.~Guan and T.~Pl{\"o}tz, ``Ensembles of deep lstm learners for activity
  recognition using wearables,'' \emph{Proceedings of the ACM on Interactive,
  Mobile, Wearable and Ubiquitous Technologies}, vol.~1, no.~2, pp. 1--28,
  2017.

\bibitem{inoue2018deep}
M.~Inoue, S.~Inoue, and T.~Nishida, ``Deep recurrent neural network for mobile
  human activity recognition with high throughput,'' \emph{Artificial Life and
  Robotics}, vol.~23, no.~2, pp. 173--185, 2018.

\bibitem{lu2019gaim}
L.~Lu, Y.~Lu, R.~Yu, H.~Di, L.~Zhang, and S.~Wang, ``Gaim: Graph attention
  interaction model for collective activity recognition,'' \emph{IEEE
  Transactions on Multimedia}, vol.~22, no.~2, pp. 524--539, 2019.

\bibitem{zhang2020temporal}
J.~Zhang, F.~Shen, X.~Xu, and H.~T. Shen, ``Temporal reasoning graph for
  activity recognition,'' \emph{IEEE Transactions on Image Processing},
  vol.~29, pp. 5491--5506, 2020.

\bibitem{wu2019learning}
J.~Wu, L.~Wang, L.~Wang, J.~Guo, and G.~Wu, ``Learning actor relation graphs
  for group activity recognition,'' in \emph{Proceedings of the IEEE Conference
  on Computer Vision and Pattern Recognition}, 2019, pp. 9964--9974.

\bibitem{singh2017graph}
D.~Singh and C.~K. Mohan, ``Graph formulation of video activities for abnormal
  activity recognition,'' \emph{Pattern Recognition}, vol.~65, pp. 265--272,
  2017.

\bibitem{tang2019coherence}
J.~Tang, X.~Shu, R.~Yan, and L.~Zhang, ``Coherence constrained graph lstm for
  group activity recognition,'' \emph{IEEE transactions on pattern analysis and
  machine intelligence}, 2019.

\bibitem{stikic2009multi}
M.~Stikic, D.~Larlus, and B.~Schiele, ``Multi-graph based semi-supervised
  learning for activity recognition,'' in \emph{2009 international symposium on
  wearable computers}.\hskip 1em plus 0.5em minus 0.4em\relax IEEE, 2009, pp.
  85--92.

\bibitem{zhou2018graph}
J.~Zhou, G.~Cui, Z.~Zhang, C.~Yang, Z.~Liu, L.~Wang, C.~Li, and M.~Sun, ``Graph
  neural networks: A review of methods and applications,'' \emph{arXiv preprint
  arXiv:1812.08434}, 2018.

\bibitem{reiss2012introducing}
A.~Reiss and D.~Stricker, ``Introducing a new benchmarked dataset for activity
  monitoring,'' in \emph{2012 16th International Symposium on Wearable
  Computers}.\hskip 1em plus 0.5em minus 0.4em\relax IEEE, 2012, pp. 108--109.

\bibitem{9010334}
G.~{Li}, M.~{Müller}, A.~{Thabet}, and B.~{Ghanem}, ``Deepgcns: Can gcns go as
  deep as cnns?'' in \emph{2019 IEEE/CVF International Conference on Computer
  Vision (ICCV)}, 2019, pp. 9266--9275.

\bibitem{DBLP:journals/corr/HeZR015}
\BIBentryALTinterwordspacing
K.~He, X.~Zhang, S.~Ren, and J.~Sun, ``Delving deep into rectifiers: Surpassing
  human-level performance on imagenet classification,'' \emph{CoRR}, vol.
  abs/1502.01852, 2015. [Online]. Available:
  \url{http://arxiv.org/abs/1502.01852}
\BIBentrySTDinterwordspacing

\bibitem{sainath2015convolutional}
T.~N. Sainath, O.~Vinyals, A.~Senior, and H.~Sak, ``Convolutional, long
  short-term memory, fully connected deep neural networks,'' in \emph{2015 IEEE
  International Conference on Acoustics, Speech and Signal Processing
  (ICASSP)}.\hskip 1em plus 0.5em minus 0.4em\relax IEEE, 2015, pp. 4580--4584.

\end{thebibliography}
\end{document}